\documentclass[letterpaper]{article} 
\usepackage[]{aaai25}  
\usepackage{times}  
\usepackage{helvet}  
\usepackage{courier}  
\usepackage[hyphens]{url}  
\usepackage{graphicx} 
\def\UrlFont{\rm}  
\usepackage{natbib}  
\usepackage{caption} 
\usepackage{algorithm}
\usepackage{algorithmic}

\usepackage{newfloat}
\usepackage{listings}
\DeclareCaptionStyle{ruled}{labelfont=normalfont,labelsep=colon,strut=off} 
\floatstyle{ruled}
\newfloat{listing}{tb}{lst}{}
\floatname{listing}{Listing}
\nocopyright 

\newcommand{\ie}{\textit{i.e.}}
\newcommand{\eg}{\textit{e.g.}}

\usepackage{xcolor}
\usepackage{booktabs}
\usepackage{amsmath}

\title{Large Language Models Are Self-Taught Reasoners: Enhancing LLM Applications via Tailored Problem-Solving Demonstrations}
\author {
    Kai Tzu-iunn Ong,
    Taeyoon Kwon,
    Jinyoung Yeo
}
\affiliations {
    Department of Artificial Intelligence, College of Computing, Yonsei University\\
    ktio89@yonsei.ac.kr
}

\usepackage{bibentry}

\begin{document}

\maketitle

\begin{abstract}
Guiding large language models with a selected set of human-authored demonstrations is a common practice for improving LLM applications. However, human effort can be costly, especially in specialized domains (\eg, clinical diagnosis), and does not guarantee optimal performance due to the potential discrepancy of target skills between selected demonstrations and real test instances.
Motivated by these, this paper explores the automatic creation of customized demonstrations, whose target skills align with the given target instance. 
We present \textbf{\textsc{Self-Taught}}, a problem-solving framework, which facilitates demonstrations that are ``\textit{\textbf{tailored}}'' to the target problem and ``\textit{\textbf{filtered}}'' for better quality (\ie, correctness) in a zero-shot manner.
In 15 tasks of multiple-choice questions of diverse domains and the diagnosis of Alzheimer's disease (AD) with real-world patients, \textsc{Self-Taught} achieves superior performance to strong baselines (\eg, Few-shot CoT, Plan-and-Solve, Auto-CoT). We conduct comprehensive analyses on \textsc{Self-Taught}, including its generalizability to existing prompting methods and different LLMs, the quality of its intermediate generation, and more.\footnote{Codes, prompts, and expert-annotated demonstrations used in our experiments are in Appendices.} 

\end{abstract}

\section{Introduction}
\label{sec:intro}
Recently, large language models (LLMs) have emerged as an alternative knowledge source to human experts, popularizing the paradigm of prompting them to solve problems.
In this context, methods such as chain-of-thought (CoT) prompting~\citep{wei2022chain}, which promotes LLMs to follow the step-by-step fashion of human problem-solving, have brought promising performances to LLM applications in diverse specialized domains~\citep{singhal2023large, zaki2024mascqa, liu2024integrating}, such as predicting crystal structures, clinical diagnosis, etc.

Despite the success, most prompt-driven projects rely on human effort. That is, they require domain experts to select representative problems for the task, annotate their solutions (\ie, rationales \& answers), and use them as demonstrations to guide LLMs in solving test instances (\ie, few-shot prompting).
Such manual effort can make real-world applications costly and, more importantly, has no guarantee of optimal performances due to the one-size-fits-all selection of problem-solving demonstrations~\citep{min-etal-2022-rethinking}, \ie, fixed and potentially unrelated (to the test instance) demonstrations used throughout the inference of the whole test set.

\begin{figure*}[th!]
    \centering
    \includegraphics[width=1\textwidth]{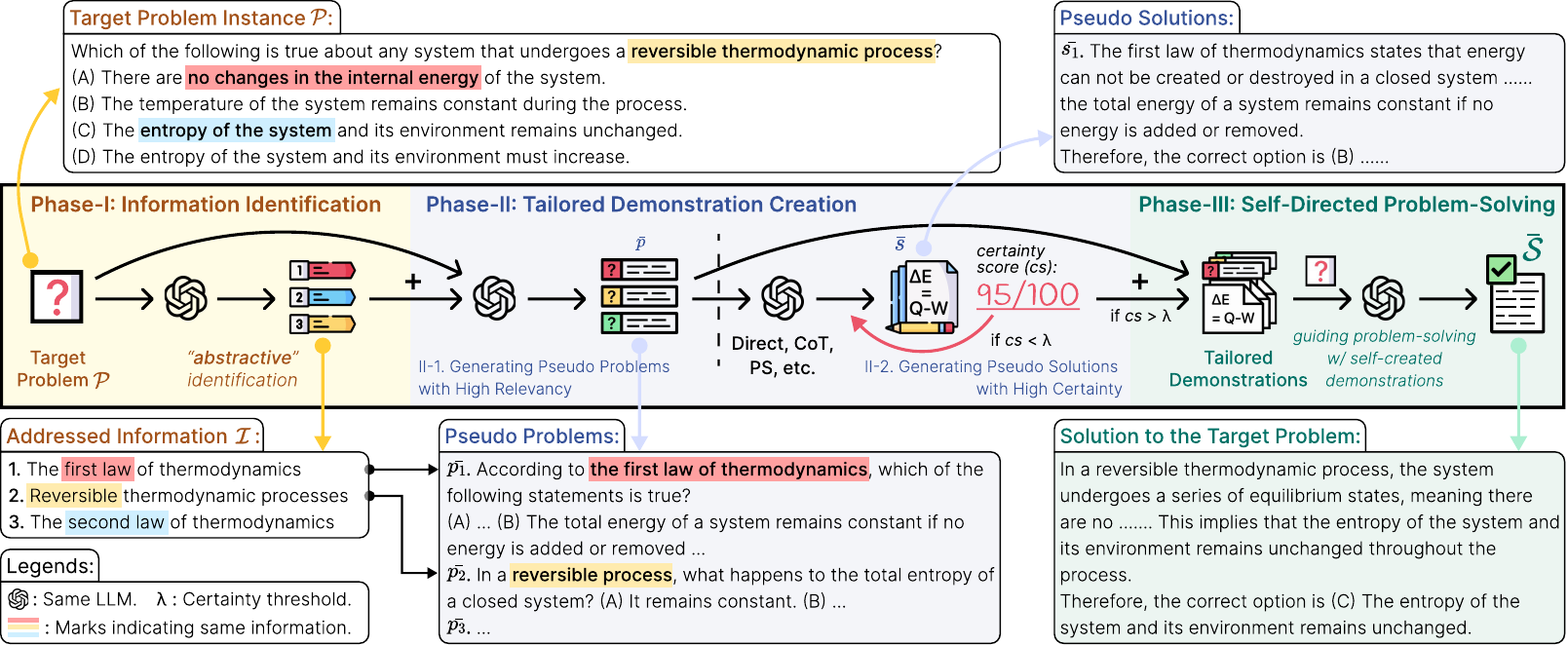}
    \caption{Empirical examples and the overview of \textbf{\textsc{Self-Taught}}. All phases are executed under a zero-shot setting.}
    \label{fig:overview}
\end{figure*}

A common remedy to such reliance on crafted demonstrations is zero-shot prompting, \ie, prompting LLMs without demonstrations. 
Upon this, studies have proposed to enhance LLMs' zero-shot reasoning~\citep{kojima2022large, chae2024language, kong2024better}. 
For instance, \citet{wang2023plan} present Plan-and-Solve (PS) prompting, where the LLM first devises a plan for the given problem and solves it according to the plan.
However, the lack of problem-solution demonstrations still sometimes poses performance gaps between zero-shot approaches and their few-shot counterparts~\citep{kojima2022large}. 
While there is a line of studies proposing to resolve this with automatic demonstration generation, they often require in-domain corpora (\ie, training/test sets) and do not explicitly address the alignment of knowledge/skills between demonstrations and test instances~\citep{zhang2022automatic, wan2023better, li2024self}.

This paper tackles the above bottlenecks in prompt-driven applications of LLMs for specialized domains.
Specifically, we focus on invoking the LLM to self-create high-quality and tailored demonstrations for each test instance under a zero-shot setting, and using them to guide its own predictions.
To this end, we present \textbf{\textsc{Self-Taught}}, a zero-shot framework of self-directed problem-solving.

Our contributions are three-fold: \textbf{(1)} We present a simple and fully zero-shot framework, \textbf{\textsc{Self-Taught}}. Inspired by self-directed learning (SDL) in educational theories,\footnote{SDL focuses on activating ones' new problem-solving ability by using their prior knowledge to reflect on related contextualized problems~\citep{christensen1991education, grow1991teaching}.}
\textsc{Self-Taught} starts by identifying information addressed in the target problem \textit{abstractively} (Phase I). After that, it goes through a tailored creation phase (Phase II), where the LLM creates problems addressing similar information/knowledge to the target as well as their solutions with \textit{high certainty}. Lastly, the self-created problems/solutions are used as tailored demonstrations for solving the target problem (Phase III);
\textbf{(2)} In 13 QA tasks of specialized domains and 2 clinical datasets collected from real-world patients of Alzheimer's disease (AD),
\textsc{Self-Taught} shows superior performances to strong baselines, including those powered by domain experts and in-domain demonstration pools;
\textbf{(3)} In our analyses, we justify \textsc{Self-Taught}'s design via ablation, show its generalizability to existing prompting methods and different LLMs, confirm the quality of our self-created demonstrations, and more.

\section{Formulations}

Many LLM-based projects~\citep{singhal2023large, liu2024integrating} use human-authored demonstrations $\mathcal{D}_{\textrm{\textit{human}}}$ to guide LLMs in generating the solution ${\mathcal{S}}$ to a given problem $ {\mathcal{P}}$:
\begin{equation}
\label{eq:disease_diagnosis}
     {\mathcal{S}} \sim P_{\theta}(\cdot| {\mathcal{P}},\mathcal{D_\textit{human}})
\end{equation}
Here, $\mathcal{D}_{\textrm{\textit{human}}}$ often contains pairs of: (1) a selected representative problem of the task; (2) a solution including intermediate reasoning steps (\ie, rationale) and a final answer.

However, the need for crafted demonstrations can make applications in specialized domains costly.
More importantly, it is not feasible to customize demonstrations for each test instance, potentially yielding sub-optimal performance due to the discrepancy between them~\citep{min-etal-2022-rethinking} -- problems used as demonstrations and the target may require completely different knowledge to solve, even if they are from the same domain. For instance, when solving physics problems, a demonstration of thermodynamics may not be beneficial to solving a problem of electronics.

Thus, we propose to create high-quality demonstrations $\mathcal{D_\textit{self}}$ via the knowledge of the LLM itself, which are tailored to each test instance.
Formally, given $\mathcal{P}$, the LLM first create $\mathcal{D_\textit{self}}$ tailored for $\mathcal{P}$, and use it to guide the prediction of $\mathcal{S}$: 
\begin{align}
\label{eq:reasoning_diagnosis}
    \mathcal{D_\textit{self}} \sim P_{\theta}(\cdot| {\mathcal{P}}) \\
     {\mathcal{S}} \sim P_{\theta}(\cdot| {\mathcal{P}}, \mathcal{D_\textit{self}})
\end{align}

\section{Proposed Framework: \textsc{Self-Taught}}
\label{sec:methods}
As shown in Figure~\ref{fig:overview}, given a target problem $\mathcal{P}$, our framework carries out the following phases in a zero-shot manner: 

\subsection{Phase I: Information Identification}
\label{ssec:phase1}
To facilitate tailored demonstrations for the target problem $\mathcal{P}$, it is important for us to first know what $\mathcal{P}$ is targeting.
Thus, \textsc{Self-Taught} starts with an information identification, where we capture what kind of knowledge/skill is addressed by $\mathcal{P}$.
Formally, given $\mathcal{P}$, the LLM lists the necessary information $\mathcal{I}$ that one must know for solving $\mathcal{P}$:
\begin{align}
\label{eq:phase1}
    \bar{\mathcal{I}} = \underset{\mathcal{I}}{\text{argmax}}\,  P_{\text{LLM}}(\mathcal{I}|\mathcal{P})
\end{align}
Note that rather than print out the information specifically in the form of factual statements (\eg, ``{\textit{According to the 2nd law of thermodynamics, idealized reversible processes produce no entropy and no process is...}''), the LLM lists the required information in an abstractive manner (\eg, ``\textit{Understanding the 2nd law of thermodynamics}''), as shown in Figure~\ref{fig:overview} bottom left.
This approach is designed to mitigate the potential influences of hallucination~\citep{lyu2022z}. We compare this method with a specific identification in Table~\ref{tab:ablation}.

\subsection{Phase II: Tailored Demonstration Creation}
\label{ssec:phase2}
Now, the LLM leverages the identified information $\mathcal{I}$ to prepare tailored problem-solution demonstrations for $\mathcal{P}$:

\subsubsection{\textit{II-1}. Generating Pseudo Problems with High Relevancy.}
We first create pseudo problems that target the same knowledge/skills as $\mathcal{P}$ based on the identified information $\mathcal{I}$. Formally, given $\mathcal{P}$ and $\bar{\mathcal{I}}$, the LLM generates a pseudo problem $\bar{p}$ targeting information listed in $\bar{\mathcal{I}}$:
\begin{align}
\label{eq:phase2}
    \bar{p} = \underset{p}{\text{argmax}}\,  P_{\text{LLM}}(p|\mathcal{P}, \mathcal{I})
\end{align}

\subsubsection{\textit{II-2}. Generating Pseudo Solutions with High Certainty.}
Intuitively, after generating $\bar{p}$, we can obtain its solution $\bar{s}$ via any zero-shot prompting approach (\eg, CoT or PS). However, since LLMs may produce non-factual statements, it is necessary to address the correctness of pseudo-solutions.

Filtering low-quality outputs with the LLM itself in a zero-shot manner is challenging.
Inspired by how LLMs can verbalize their confidence in their predictions,\footnote{In math reasoning tasks, \citet{xiong2024can} find LLMs predictions to be more often correct when expressed a higher certainty.} we apply a \textbf{Certainty Filtering}.
Formally, given a pseudo problem $\bar{p}$, the LLM first create a pseudo solution $\bar{s} = (\bar{r}, \bar{a})$ consisting of a rationale $\bar{r}$ and an answer $\bar{a}$. Next, it outputs a certainty score $\bar{cs}$ within 0-100 before ending the generation. The form of $\bar{r}$ depends on the zero-shot prompting method of one's choice, \eg, \textbf{\texttt{Direct Prediction}}, \textbf{\texttt{CoT}}, \textbf{\texttt{PS}}, etc:\footnote{We use zero-shot CoT~\citep{kojima2022large} when not specified.}
\label{ssec:phase3}
\begin{align}
\label{eq:phase3}
    \bar{s} = \underset{s}{\text{argmax}}\,  P_{\text{LLM}}(s|\bar{p})\\
    \Rightarrow \bar{cs} = \underset{cs}{\text{argmax}}\,  P_{\text{LLM}}(cs|\bar{p}, \bar{s})
\end{align}
where $\Rightarrow$ indicates the sequential generation of tokens. 
To collect highly-confident $s$, we iterate this process (for at most $t$ times) until we get a $\bar{s}$ to $\bar{p}$ that yields a ${cs} \geq \lambda$.

In practice, we collect plural pairs (\ie, $N$ pairs)\footnote{We empirically set $N$ to 3, $\lambda$ to 90, and $t$ to 5. In practice, we randomly select an $s$ with the highest $cs$ if we cannot obtain any $s$ with $cs \geq \lambda$ before the end of the $t$-th iteration.} of $\bar{p}$ and $\bar{s}$ to build a set of self-created tailored demonstrations $\mathcal{D}_{\textrm{\textit{self}}}$:
\begin{align}
\label{eq:phase3}
    \mathcal{D}_{\textrm{\textit{self}}} = \{(\bar{p_n}, \bar{s_n})\}_{n=1}^N
\end{align}

\subsection{Phase III: Self-Directed Problem-Solving with Tailored Demonstrations}
While many works emphasize diverse and representative demonstrations of the test set~\citep{zhang2022automatic, singhal2023large, li2024self}, it may lead to unrelated demonstrations that provide sub-optimal or misleading guidance, eventually affecting the solving process of target problems~\citep{lightman2024let}, especially in scenarios where each test instance requires different knowledge to solve.

We address this by guiding LLMs' problem-solving with its self-created demonstrations, which are tailored to the target.
Formally, given $\mathcal{P}$, $\mathcal{D}_{\textrm{\textit{self}}} = \{\bar{p_n}, \bar{s_n}\}_{n=1}^N$ is added to the LLM's input to guide the prediction of the solution $\bar{\mathcal{S}}$ to $\mathcal{P}$:
\label{ssec:phase4}
\begin{align}
\label{eq:phase4}
    \bar{\mathcal{S}} = \underset{\mathcal{S}}{\text{argmax}}\,  P_{\text{LLM}}(\mathcal{S}|\mathcal{P}, \mathcal{K}_{\textrm{\textit{self}}})
\end{align}

\section{Datasets}
To investigate the effectiveness of ours,
we adopt several tasks from specialized domains, starting with multiple-choice questions of different emphases:

\subsection{Question Answering of Diverse Domains}
\textbf{StrategyQA.} This dataset contains questions that target multi-hop reasoning over a wide range of knowledge~\citep{geva2021did}.
For example, a question may require one to know facts about a celebrity and the properties of hydrogen.
\newline \textbf{ScienceQA.} Proposed by \citet{lu2022learn}, questions are collected from elementary $\sim$ high school curricula, targeting 26 topics including math, language, geography, etc. We exclude problems addressing the vision 
modality, \ie, images.
\newline \textbf{MedQA.} A popular benchmark datasets in the medical domain curated by \citet{jin2021disease}. The questions are collected from national medical licensing exams in several countries. We only adopt questions written in English.
\newline \textbf{College-level problems of six domains.} We include college-level problems of computer science (CS), medicine (Med), chemistry (Chem), math, physics (Phys), and biology (Bio) domains. The problems are collected from college textbooks and exams by \citet{hendrycks2020measuring}.
\newline \textbf{Professional-level problems of four domains.} We include four datasets addressing professional-level knowledge of accounting (Acct), medicine (Med), psychology (Psych), and legal (Law) domains. The problems are mainly obtained from diverse licensing/bar exams of the corresponding profession by \citet{hendrycks2020measuring}.

\subsection{Clinical Diagnosis with Real-World Patients}
Besides problems from academic scenarios, we further evaluate \textsc{Self-Taught} on a long-standing real-world challenge: the diagnosis of Alzheimer's disease (AD).
For that, we incorporate two datasets of actual patients, collected by \textbf{ADNI}~\citep{jack2008alzheimer} and \textbf{AIBL}~\citep{ellis2009australian}.

AD diagnosis requires the LLM to reason over the electronic health records (EHRs) of patients and make the diagnosis accordingly, \ie, either \texttt{AD}, \texttt{MCI} (mild cognitive impairment), or \texttt{Normal}.
The EHRs are structured following the practice of/obtained from \citet{kwon2024large}, which list the findings from MRI scans (\eg, volume measurements of each brain region) and patient information such as the results of mental state exams, the presence of APOE4 allele, etc. Details and examples of all datasets are in Appendices.

\begin{table*}[t!]
\setlength{\tabcolsep}{3pt}
\centering
\small
\begin{tabular}{lccccccccccccc||c}
\toprule
& & & & \multicolumn{6}{c}{\textbf{MMLU: College Level}} & \multicolumn{4}{c}{\textbf{MMLU: Professional Level}} & \\
\cmidrule(lr){5-10} \cmidrule(lr){11-14}
\textbf{Methods / Tasks}
 & StrategyQA & ScienceQA  & MedQA & CS & Med & Chem & Math & Phys & Bio & Acct & Med & Psych & Law & \textbf{Avg} \\ 
 \midrule
\multicolumn{5}{l}{\textit{Zero-shot prompting / without real demonstrations:}}  \\ 
\midrule
Zero-shot Direct& 66.11&82.42& 56.64 & 55.00&64.16& 38.00 & 31.00 & 45.10 & 65.28 & 44.68 & 76.84 &  64.38 & 46.35 & 56.61 \\

Zero-shot CoT & 70.96 & 87.59 & 68.11 & 61.00 &  71.10 & 52.00 & 43.00 & 59.80 & 79.86 & 58.51 & \textbf{81.25} & 72.22 & 50.13 &  65.81 \\
Plan-and-Solve & 70.39&87.72&66.93 & \textbf{65.00} &69.94& 54.00 &\textbf{49.00}&54.90&79.17 & \textbf{60.28}  & \textbf{81.25} & 71.24 & 49.87 & 66.13 \\
RiC prompting & 64.63 & 87.63 & 57.66 & 54.00 & 66.47 & 40.00 & 43.00 & 53.92 & 74.31 & 53.55 & 75.00 & 51.93 & 46.15 & 59.10 \\
Role-Play& 68.17 & 86.83 & 60.49 & 60.00 & 72.25 & 51.00 & 45.00 & 59.80 & 79.86 & 57.09 & 75.00 &  70.75 & \textbf{50.46} & 64.53 \\
\underline{\textbf{\textsc{Self-Taught}}}& \textbf{73.93} & \textbf{88.44} & \textbf{68.50} & 64.00 & \textbf{76.16} & \textbf{56.00} & 48.00 & \textbf{65.69} & \textbf{81.94} & 60.00 & \textbf{81.25} & \textbf{75.53} & \textbf{50.46} & \textbf{68.47}  \\

\midrule
\midrule
\multicolumn{8}{l}{\textit{Oracles (demonstrations are made with real problems):}}
  \\
\midrule
Few-shot Direct & \underline{66.99} & \underline{86.20} & \underline{57.71} & \underline{54.64} &  \underline{68.24} & \underline{42.71} & \underline{37.23} & \underline{50.00} & \underline{80.14} & \underline{51.25} & \underline{79.18} &  \underline{73.89} & 51.08 & \underline{61.48} \\

Manual CoT & 74.81 & \underline{87.37} & 69.99 & \underline{63.92} &  \underline{73.41} & \underline{55.21} & 50.00 & \underline{62.50} & \underline{78.01} & \underline{58.06} & \underline{80.51} &  \underline{75.04} & \underline{45.11} & \underline{67.19}\\
Retrieval CoT& \underline{70.31} & 89.43 & 69.44 & 65.00 &  \underline{71.68} & 56.00 & 49.00 & \underline{56.86} & \underline{79.86} & \underline{54.61} & 82.72 &  \underline{72.50} & \underline{48.71} & \underline{66.58} \\
Auto-CoT & \underline{73.11} & \underline{88.17} & 70.93 & \underline{57.45} & \underline{70.00} &\underline{55.21} & \underline{45.74} & \underline{54.64} & \underline{78.72} & \underline{58.42} & 81.78 &  \underline{75.37} & \underline{49.05} & \underline{66.05} \\

\bottomrule
\end{tabular}

\caption{Model performances (accuracy) in question-answering. \underline{Underlines}: Oracles that are outperformed by ours.}
\label{tab:main_results_QA}
\end{table*}

\section{Experimental Settings}

\subsection{Zero-shot Baselines}
Zero-shot prompting has been widely utilized to mitigate human effort in LLM applications.
Since \textsc{Self-Taught} also access to nothing but the target $\mathcal{P}$ and the LLM's own knowledge,
we consider these fair comparisons:
\newline \textbf{Direct prediction (Direct).} The LLM directly predicts the solution for $\mathcal{P}$ without any intermediate process.
\newline \textbf{Chain-of-Thought prompting (CoT).} Given a target problem $\mathcal{P}$, this setting promotes the LLM to generate the intermediate reasoning steps towards the solution using the phrase ``\textit{Let's think step-by-step}''~\citep{kojima2022large}.
\newline \textbf{Plan-and-Solve prompting (PS).} \citet{wang2023plan} present PS prompting, which has shown promising performance in solving math problems. In PS, the LLM first devises a plan based on the problem instance and then solve it step-by-step according to the plan.
\newline \textbf{Reasoning-in-Conversation (RiC).} Inspired by its performance in linguistic tasks such as humor detection~\citep{wang2024reasoning}, we modified it for our experiments. This setting prompts an LLM to first simulate a discussion between experts and then conclude the answer from the conversation. 
\newline \textbf{Role-Play prompting.} Proposed by \citet{kong2024better}, it outperforms zero-shot CoT in commonsense and math reasoning tasks.
Here, the LLM and the user kick off the session with a short conversation that helps the LLM get into the role of an expert (\eg, a math teacher). Then, the problem-solving will be performed in a ``teaching the user'' manner.

\subsection{Few-shot Baselines (Oracles)}
We further compare \textsc{Self-Taught} with few-shot prompting methods. 
Since these methods have access to demonstrations made with real problems from the dataset, we consider them oracles following \citet{lyu2022z}:
\newline \textbf{Manual Chain-of-Thought (Manual 
     CoT).} A popular setting of few-shot CoT prompting in LLM applications, where the problem-solving is guided by demonstrations written by human domain experts (curated by prior work or our domain experts). Details are provided in Appendices.
\newline \textbf{Retrieval CoT.} Following \citet{zhang2022automatic}, when given $\mathcal{P}$, we retrieve top-$N$ similar problems from the training set via text similarity. 
Then, we use the LLM to annotate CoT rationales/solutions for the retrieved problems, using them as demonstrations for solving $\mathcal{P}$.
Similar to \textsc{Self-Taught}, this setting also pursues demonstrations that address similar or identical knowledge/information to $\mathcal{P}$.
\newline \textbf{Automatic CoT (Auto-CoT).} It is widely applied for scenarios where demonstrations are unavailable~\citep{zhang2022automatic}. It is similar to Retrieval-CoT, but we instead sample $N$ most ``diverse'' problems from the training set via k-clustering as demonstrations that represent the whole task.

\subsection{Models and Implementation Details} 
\textbf{Large language models.} We use gpt-3.5-turbo-0125 and llama-3.1-8B~\citep{openai2023chatgpt, meta2024llama}, w/ temperature of $0.7$. We report GPT's results in Table 1-3, Llama's summarized results in RQ6, and full results in Appendices.
\newline \textbf{Encoder for Auto-/Retrieval CoT.} Following \citet{zhang2022automatic}, we use Sentence-BERT~\citep{reimers2019sentence} to encode problems for clustering and retrieval.
\newline \textbf{Demonstrations in few-shot baselines.} If a dataset does not provide a training set, the demonstrations will be based on instances from the test set.
They will be ignored during performance measurements. Also, we set $N$ to 3 for QA. For AD diagnosis, we set it to 2 to match the radiologist demonstrations provided by \citet{kwon2024large}.
\newline \textbf{Evaluation Metrics.} We report accuracy (\%). For AD diagnosis, we further include precision, recall, and F1 score for a more comprehensive comparison.
\newline \textbf{Preventing randomness.} We report the median performance of three runs for all experiments in this work.

\section{Results and Discussions}
We present the results of the following Research Questions:
\newline \textbf{RQ1:} Can \textsc{Self-Taught}'s tailored demonstrations enhance the LLM's reasoning in QA of diverse domains?
\newline \textbf{RQ2:} Is \textsc{Self-Taught} also beneficial in the clinical diagnosis with real-world patients of Alzheimer's disease?
\newline \textbf{RQ3:} How phases in \textsc{Self-Taught} affect performances?
\newline \textbf{RQ4:} Can \textsc{Self-Taught} also be applied to other zero-shot prompting methods besides CoT, and vice versa?
\newline \textbf{RQ5:} How cost-efficient is \textsc{Self-Taught}?
\newline \textbf{RQ6:} Can \textsc{Self-Taught} generalize to other LLMs?

\subsubsection{Tailored demonstrations allow \textsc{Self-Taught} to outperform baselines in diverse QA tasks (RQ1).}
Table~\ref{tab:main_results_QA} shows model performances in 13 QA tasks. \textsc{Self-Taught} ranks first in 10 tasks, second in the rest of 3, and achieves better average acc than zero-shot baselines (top half).

Compared with oracles, ours outperforms Manual and Auto-CoT in 10 and 11 tasks (out of 13).
This suggests that using a fixed set of demonstrations (whether human-written or machine-generated) throughout the inference of all test data may yield sub-optimal performance, justifying our goal of tailored demonstrations.
Retrieval CoT performs almost as well as ours when compared head-to-head (6 vs. 7 wins). We assume that it is because it also leverages problems relevant to $\mathcal{P}$ as demonstrations.
Still, ours yields a much higher Avg acc, indicating that our generative method can elicit better related demonstrations than similarity-based retrieval.

In Law, few-shot Direct outperforms all settings that involve intermediate reasoning (\ie, rationale). This may be because while problems are collected from the US, there is no clear regulation specifying which state's law should be referenced in each problem. This can cause misquotation of law and make the correct annotation/generation of rationales challenging, negatively affecting problem-solving.
Interestingly, in Physics (Phys), \textsc{Self-Taught} and Manual CoT have much higher acc than others. This may suggest that ensuring the quality of demonstrations is relatively more important in physics domains than in other domains.\footnote{When the certainty filtering in \textsc{Self-Taught} is ablated, the accuracy drops by $2.94$ percentage points to $62.75$.}

\begin{table}[t!]
\centering
\setlength{\tabcolsep}{4pt}
\small
\begin{tabular}{lcccc}
\toprule
\textbf{Patients from:} &\multicolumn{4}{c}{\textbf{ADNI \& AIBL}} \\
\midrule
      \textbf{Methods / Metrics} & \textbf{F1} & \textbf{Precision} & \textbf{Recall} & \textbf{Accuracy} \\ 
\midrule 
Zero-shot Direct & 49.22 & 52.90 & 53.54 & 53.54 \\
Zero-shot CoT & 52.81 & 54.92 & 55.97 & 55.97 \\
Plan-and-Solve & 50.57 & 53.49 & 55.70 & 55.88 \\
RiC Prompting & 51.12 & 51.16 & 52.54 & 52.54 \\
Role-Play & 48.73 & 55.62 & 56.49 & 56.49 \\
\textbf{\textsc{Self-Taught}} & \textbf{56.08} & \textbf{58.15} & \textbf{59.27} & \textbf{58.56} \\
\midrule \midrule
\multicolumn{5}{l}{\textit{Oracles (demonstrations are made with real problems):}}
\\ 
\midrule
Few-shot Direct & \underline{44.03} & \underline{52.10} & \underline{54.47} & \underline{54.47} \\
Manual CoT & 63.12 & 64.19 & 64.72 & 64.73 \\
Retrieval CoT & \underline{50.32} & \underline{54.49} & \underline{52.25} & \underline{52.20} \\
Auto-CoT & 56.27 & \underline{57.31} & \underline{57.75} & \underline{57.75} \\

\bottomrule
\end{tabular}
\caption{Model performances in AD diagnosis (average of datasets). \underline{Underlines}: Oracles outperformed by ours. Precision, recall, and F1 are weighted Avg of 3 diagnosis classes.}
\label{tab:main_results_AD}
\end{table}

\subsubsection{Ours is less effective than Manual CoT in AD diagnosis due to the high similarity between instances (RQ2).}
In AD diagnosis (Table~\ref{tab:main_results_AD}), \textsc{Self-Taught} beats all baselines in all metrics, except for Manual CoT.
This pattern is much different from the findings in RQ1.
We conjecture that it is because the discrepancy between each instance is extremely small here: all instances require LLMs to perform the same task w/ the same 3 output classes, via EHRs that are structured in the identical key-value format.
This can marginalize the effect of tailored demonstrations and amplify the benefit of fixed human-crafted demonstrations.

Regardless, ours presents a much smaller performance gap between it and Manual-CoT ($6.18$ percentage points of acc; avg of all classes/datasets) than other baselines ($8.25 \sim 12.54$ percentage points).
In real clinical settings, each patient's EHR may be constructed diversely due to each radiologist's preference and other situational factors, thus requiring different reasoning to diagnose~\citep{norman2005research}.
With insights from RQ2, we presume one can adjust \textsc{Self-Taught} for real-world applications by combining it with slight manual effort. For instance, including a minimal demonstration in the generation of pseudo problems/EHRs (Phase II), \ie, demonstration expansion, to tackle different EHR styles and diagnostic processes. We leave this to future work.


\begin{table*}[ht!]
\setlength{\tabcolsep}{3pt}
\centering
\small
\begin{tabular}{lccccc|cc||c}
\toprule
\textbf{Settings / Tasks}
 & StrategyQA & ScienceQA  & MedQA & COLLEGE & PRO & ADNI & AIBL & \textbf{Avg}\\ 
\midrule
\textbf{\textsc{Self-Taught}} (Ours; as reference) & 73.93 & 88.44 & 68.5 & 67.42 & 60.33 & 60.34 & 56.78 & 67.96 \\\cmidrule(lr){1-1} \cmidrule(lr){2-9}

w/o Information Identification (Phase I) & \underline{73.10} & 87.95 & \underline{63.71} & 61.79 & 57.08 & 59.29 & 53.97 & 65.27 (-2.69) \\

w/o Certainty Filtering (CF; in Phase II-2) & \underline{73.10}	& 86.87 & 65.36 & 66.48  & 60.15 & 59.95  & 56.54 & 66.92 (-1.04) \\
w/o Both & \underline{73.10} & 86.87 & 65.36 & \underline{61.63} & 56.84 & \underline{58.89} & \underline{53.04}  & \underline{65.10} (-2.86) \\
Specific Information Identification & 74.24 & \underline{86.74} & 64.65 & 62.90 & \underline{56.16} & 59.82 & 54.67 & 65.60 (-2.36) \\

\bottomrule
\end{tabular}
\caption{Performance (accuracy) of ours' ablations and modification. COLLEGE and PRO are the weighted Avg of the corresponding 6 and 4 datasets. \underline{Underlines} are the worst performances across variants of \textsc{Self-Taught}.}
\label{tab:ablation}
\end{table*}

\subsubsection{Designed phases contribute to performance improvement (RQ3).}
To investigate how our phasic design affects model performance, we evaluate ablated and modified \textsc{Self-Taught} in all 15 tasks in
Table~\ref{tab:ablation}.
First, ablations of information identification and certainty filtering both lead to worse performance, while the former has a larger impact.\footnote{We prompt the LLM to generate pseudo questions that address the same/similar information as $\mathcal{P}$ directly based on $\mathcal{P}$ only.}
This shows: (1) having a phase for identifying what $\mathcal{P}$ is targeting is crucial for creating tailored demonstrations, which is beneficial for CoT prompting; (2) when tailored pseudo problems are available, quality-controlling their solutions can further boost system performance.

We also report a version where the information identification (Phase I) is done by printing out the specific factual statements that are required to solve $\mathcal{P}$.
We find this performing worse than the \textsc{Self-Taught} and the 2nd ablation (both are equipped with an abstractive identification). This justifies our design of Phase I.


\subsubsection{\textsc{Self-Taught} enhances existing prompting methods (RQ4).}
So far, all solution generation of pseudo and target problems (\ie, Phase II-2 and III) has been driven by CoT prompting. As a formal study on improving LLM applications, it is necessary to validate \textsc{Self-Taught}'s efficacy when implemented with different prompting methods.
Thus, we repeat the above experiments with \textsc{Self-Taught}'s variants, where the generation of solutions (Phase II-2 and III) is based on zero-shot direct and PS prompting.

Figure~\ref{tab:improvement} reports the improvement of average acc in all 15 tasks.\footnote{The dataset-specific results are available in Appendices.}
Firstly, when incorporating \textsc{Self-Taught} to existing methods, we observe performance gains in all of them. The most significant improvement is in 0-shot Direct, where system performance increases by 2.35 percentage points of acc.
Zero-shot PS benefits not as much from \textsc{Self-Taught}, we hypothesize that it is because the ``plan'' devised in PS has already worked as a self-created guidance for problem-solving, marginalizing the help of self-generated demonstrations from \textsc{Self-Taught}.
We also present the improvement brought by \textsc{Self-Taught} w/o certainty filtering (CF). Performance gains are still achieved. This provides us with a relatively more cost-efficient implementation (than original \textsc{Self-Taught}) for incorporating existing methods.

\begin{table}[t!]
\centering
\setlength{\tabcolsep}{4pt}
\small
\begin{tabular}{lccc}
\toprule
\textbf{Method:} & Direct & CoT & PS \\ 
\midrule 
None & 59.68 & 65.86 & 65.56\\\midrule
\textbf{\textsc{Self-Taught}} & 62.03 (+2.35) & 67.96 (+2.10) & 66.68 (+1.12)\\
w/o CF & 61.21 (+1.53) & 66.92 (+1.06) & 65.87 (+0.31)\\
\bottomrule
\end{tabular}
\caption{Performance (avg. accuracy of all 15 datasets) when powering \textsc{Self-Taught} with diverse zero-shot prompting.}
\label{tab:improvement}
\end{table}

\subsubsection{There exists a cost-performance trade-off (RQ5).}

A concern of \textsc{Self-Taught} is its higher API cost. Regardless, we argue that ours is competitive when taking both performance and cost into account. Figure~\ref{fig:cost} plots accuracy against gpt-3.5-turbo-0125's API cost per instance (calculated based on input and output tokens in six college-level tasks). We find \textsc{Self-Taught} and its ablation (w/ certainty filtering) lying on the Pareto frontier, indicating an efficient cost-performance trade-off. This suggests \textsc{Self-Taught}'s value when performance is prioritized over the API cost.

\begin{figure}[h!]
    \centering
    \includegraphics[width=0.8\linewidth]{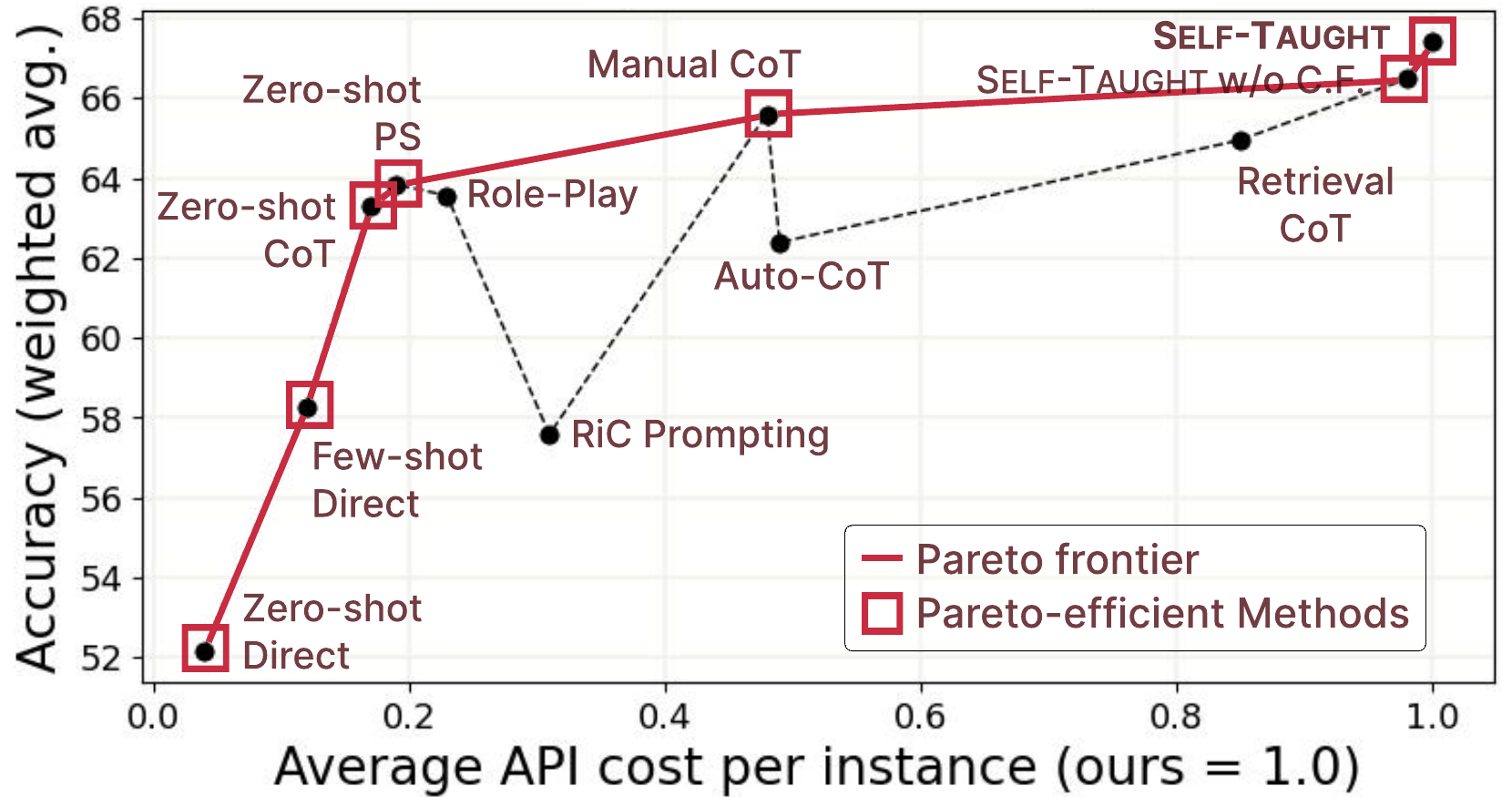}
    \caption{Cost-performance comparisons (ours' cost = $1.0$).}
    \label{fig:cost}
\end{figure}

\subsubsection{\textsc{Self-Taught} generalizes to a smaller open-source LLM (RQ6).}
To address RQ5, we test if ours' can generalize to an LLM that is both smaller and open-source. We report the performance of Llama-3.1-8B regarding the above Pareto-efficient methods in Figure~\ref{fig:sllm}.
\textsc{Self-Taught} generally yield the best performance among the Pareto-efficient methods, suggesting that it can be a strong candidate method in settings with limited computational power and budget.

\begin{figure}[t!]
    \centering
    \includegraphics[width=1\linewidth]{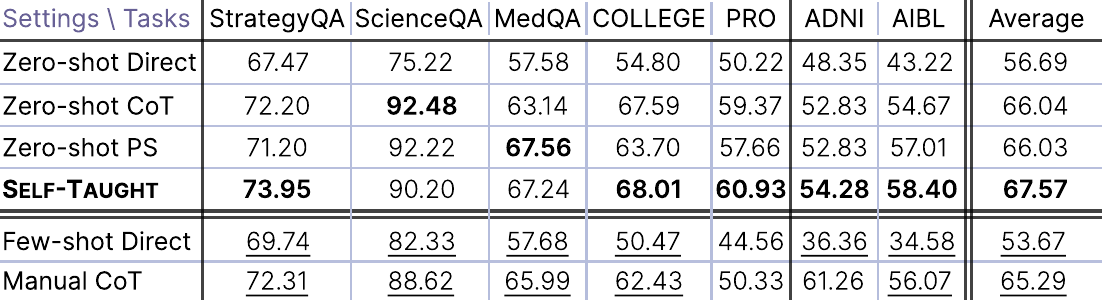}
    \caption{Ours' performances with Llama-3.1-8B (acc).}
    \label{fig:sllm}
\end{figure}

\section{Human Evaluations and Further Analyses}

\subsection{Evaluating \textsc{Self-Taught}'s Intermediate Phases}
To further assess \textsc{Self-Taught}'s intermediate phases, we conduct human evaluations on 108 pseudo problem/solution pairs sampled from six college-level tasks (Figure~\ref{fig:human}).

Overall (red numbers), thanks to proper information identification (94.4\%), 86.1\% of the pseudo problems address similar information as target $\mathcal{P}$. Also, 83.3\% of pseudo solutions contain helpful CoT rationales and 77.8\% of them correctly answer the pseudo problems. These suggest the quality of \textsc{Self-Taught}'s intermediate generation.

When looking at \textsc{Self-Taught}'s correct and wrong problem-solving separately. The largest difference (blue numbers) is that successful problem-solving is usually in the company of helpful CoT rationales (94.5\%) and correct final answers (91.7\%) in the pseudo solutions, while failed cases yield a much lower occurrence of them (58.3\% and 58.3\%).
Interestingly, failed problem-solving comes with pseudo problems that are easier than the target problem two times more often than successful cases (36.1\% vs. 16.7\%; green numbers). This provides a direction for future work, where one can more explicitly address the difficulty of machine-generated demonstrations to improve the final problem-solving.

\begin{figure}[h!]
    \centering
    \includegraphics[width=1\linewidth]{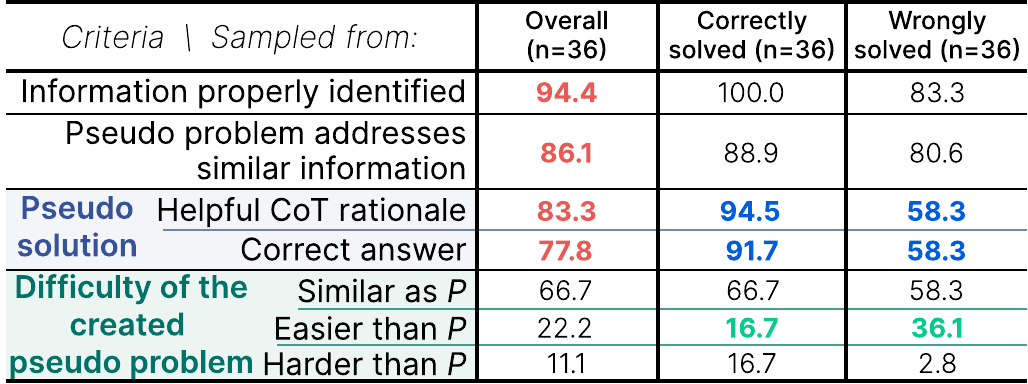}
    \caption{Human evaluation of ours' intermediate outputs. We present the percentage of approval voting.}
    \label{fig:human}
\end{figure}

\subsection{Case Study} Figure~\ref{fig:case} shows a representative case of how \textsc{Self-Taught} better elicits demonstrations that address the same knowledge as the target problem instance. In settings where demonstrations is built upon the in-domain training or test set (\eg, Retrieval CoT), unrelated demonstrations (regarding ``collision'' and ``spring'') may appear when there is no any instance in the given dataset targeting the same knowledge as $\mathcal{P}$ (targeting ``energy dissipation in the circuit''). By contrast, ours can get rid of such bottleneck and generatively facilitate tailored demonstrations with our designed phases. More empirical examples are available in Appendices.
\begin{figure}[h!]
    \centering
    \includegraphics[width=1\linewidth]{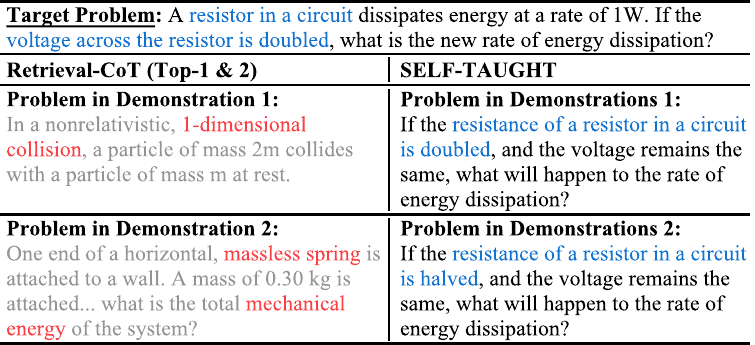}
    \caption{Demonstrations in Retrieval CoT and Ours.}
    \label{fig:case}
\end{figure}

\subsection{Comparing \textsc{Self-Taught} with Zero-shot CoT}
Here, we investigate the improvement brought by \textsc{Self-Taught} (using 0-shot CoT in Phase II and III) over vanilla CoT in Figure~\ref{fig:compare} by comparing their prediction correctness. Among problems that 0-shot CoT is wrong, \textsc{Self-Taught} solve 15.4\% of them correctly, even though they are both based on 0-shot CoT prompting. This exhibits the benefit of tailored demonstrations in LLM problem-solving.
\begin{figure}[h!]
    \centering
    \includegraphics[width=0.80\linewidth]{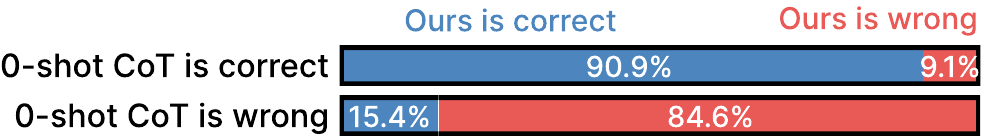}
    \caption{\textsc{Self-Taught}'s predictions in college-level problems that are correctly and wrongly solved by zero-shot CoT.}
    \label{fig:compare}
\end{figure}

\subsection{Number of Pseudo Shots}
We analyze the effect of varying the number of
self-created pseudo shots using the six college-level tasks (Figure~\ref{fig:shot}). First, \textsc{Self-Taught} outperforms Manual CoT and Retrieval CoT (both have $N$ = 3) with fewer shots ($N$ = 2).
This suggests the effectiveness of tailored demonstrations generated with our framework, as well as the possibility of tuning $N$ to further decrease our computational cost.
Also, we observe that when $N \geq 3$, model performance remains almost consistent. This pattern matches the findings from regular prompting with real demonstrations~\citep{brown2020language}.

\begin{figure}[h!]
    \centering
    \includegraphics[width=0.9\linewidth]{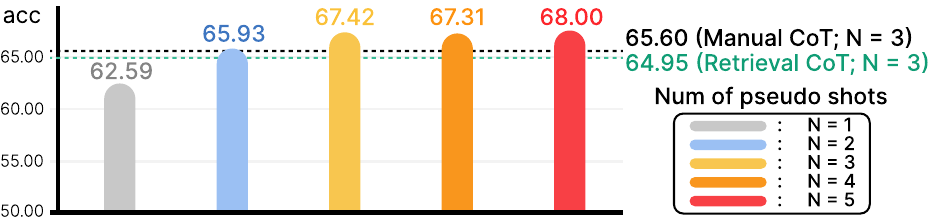}
    \caption{Ours with varying numbers of pseudo shots.}
    \label{fig:shot}
\end{figure}

\subsection{\textsc{Self-Taught} and Task Difficulty}
Lastly, we investigate the relation between \textsc{Self-Taught} and task difficulty (approximated by the most un-engineered setting, \ie, 0-shot Direct). Figure \ref{fig:regression} shows a trend that as the performance of 0-shot direct decreases, the improvement increases, indicating that the LLM benefits more from \textsc{Self-Taught} in tasks that are initially harder for it w/o additional techniques. This provides a guideline for us to judge the priority when applying ours to LLM applications. Also, when plotting CoT against Direct, the regression coefficient $\beta$ = $-0.14$, showing that \textsc{Self-Taught} generally brings more improvement ($\beta$ = $-0.21$) than CoT as the task gets difficult.

\begin{figure}[bh!]
    \centering
    \includegraphics[width=0.9\linewidth]{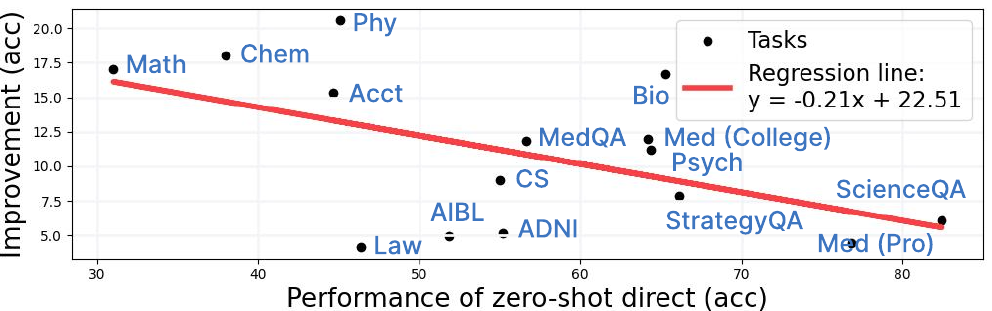}
    \caption{\textsc{Self-Taught}'s effect w.r.t. task difficulty.}
    \label{fig:regression}
\end{figure}

\section{Related Work}
General-purpose LLMs have been widely applied to diverse domains thanks to their ability to learn from user input: \citet{li2024cancergpt} prompt LLMs to predict drug synergy; \citet{chiang2024llamp} build a prompt-driven system with material science APIs for crystal generation; \citet{kwon2024large} use LLMs to annotate raw patient data. However, they often rely on human demonstrations, which can be costly and yield sub-optimal performance when encountering the above-discussed demonstration-target discrepancy.

Besides the mentioned works (\eg, PS, Auto-CoT, etc.), several studies have proposed to mitigate human efforts in prompt construction:
\citet{zhou2024self} and \citet{chae2024language} invoke LLMs to generate task-level plans shared across all instances to guide instance-level inference.
\citet{lyu2022z} use sentences from external related corpora along with random labels as pseudo demonstrations for text classification.
\citet{wan2023better} use LLMs to create a large demonstration pool by repeatedly running the test set with 0-shot CoT and using majority-vote-based criteria to select good demonstrations for problem-solving.
Similarly, \citet{li2024self} use LLMs to generate a pseudo dataset of 5K questions with 29 designed topics from scratch and use it as the demonstration pool.
Recently, \citet{yang2024large} use LLMs as optimizers, where the LLM iteratively updates the problem-solving instruction (\eg, ``Break this down'') until it yields a maximum accuracy on the training set.
To address the lack of annotated rationales for CoT fine-tuning, \citet{hwang2024self} first run the training set with LLMs to collect correct/wrong CoT rationales. Then, the ``first wrong step'' in wrong rationales is identified with designed algorithms and used as fine-grained rewards for preference learning.

Similar to ours, \citet{kim2022self} and \citet{chen2023self} generate pseudo demonstrations by prompting LMs with phrases like ``generate a negative review'' or ``Come up with diverse creative instances for the task''. However, the former requires the same output span (\eg, positive \& negative) across all test instances. The latter focuses on facilitating diverse representative problems following Auto-CoT~\citep{zhang2022automatic}. It neglects the discrepancy (addressed information) between test and pseudo problems as well as the correctness of pseudo solutions (we show the effect of such neglection in Table~\ref{tab:ablation}).
Inspired by them, \textsc{Self-Taught} resolve human effort by creating demonstrations that are both ``quality-controlled'' and ``tailored'' to each target instance, which is under-explored so far to the best of our knowledge.

\section{Conclusions}
We present \textbf{\textsc{Self-Taught}}, a problem-solving framework for specialized domains. It addresses the costly human effort and the demonstration-target discrepancy in LLM applications, by creating demonstrations that are quality-controlled and tailored to each test instance.
It outperforms baselines in 15 tasks of QA and AD diagnosis. It is Pareto efficient regarding cost and performance and generalizable to different prompting methods and LLMs. The quality of self-created problems/solutions is confirmed in expert evaluations. We discuss the limitations of our work in Appendices.

\section{Acknowledgements}
\textit{Jinyoung Yeo} is the corresponding author.
This work is partly supported by an IITP grant funded by the Korean Government
(MSIT) (No. RS-2020-II201361, Artificial Intelligence Graduate School Program of Yonsei
University).

Additionally, besides the first author, who has a BSc in mechanical engineering, the following domain experts contribute to the annotation of human demonstrations (for tasks from MMLU) as well as human evaluations:
\textit{Cheng-Wei Hsu} (licensed doctor at MacKay Memorial Hospital, Taiwan),
\textit{Chun-Hsiang Chou} (licensed doctor at Shuang Ho Hospital, Taiwan),
\textit{Shuan Chen} (PhD in chemical engineering),
\textit{William Jackson} (licensed lawyer),
\textit{Yongho Song} (MSc in computer science),
\textit{Ruo-qiao Wen} (licensed psychologist at NTU Hospital, Taiwan),
\textit{Issac Yoo} (licensed tax accountant),
\textit{Sekwon Oh} (BSc in electrical and electronic engineering), \textit{SeongHyeon Bae} (BSc in mathematics), and \textit{Bryan Oh} (BSc in electrical engineering).  We sincerely appreciate their help.

\bibliography{custom}
\clearpage
\newpage

\section{Appendices}
\setcounter{secnumdepth}{1}
\section{Limitations}
This work has limitations. First of all, \textsc{Self-Taught} can be less cost-efficient. Although we have shown that \textsc{Self-Taught}'s cost-performance trade-off is acceptable (\ie, Pareto efficient), addressing the system cost more explicitly can be necessary for real-world deployment.
A plausible direction is combining Retrieval CoT, \textsc{Self-Taught}, and an additional logic (\eg, a threshold of text similarity) that determines whether problems addressing similar knowledge to the target problem are available in the training/test corpora (for us to retrieve) and selectively generate tailored demonstrations only when such problems are absent or not enough for the desired number of shots.

Another concern is that although \textsc{Self-Taught} generally brings more performance gains to the adopted LLM in tasks that are initially more challenging for it (Figure~\ref{fig:regression}), the final performance may still be far from optimal due to the lack of related knowledge in its parameter.
One may address this by (1) adopting LLMs that are fine-tuned with corpora of the corresponding domains, \eg, running \textsc{Self-Taught} with BioMistral~\citep{labrak2024biomistral} when solving medical problems, or (2) retrieving relevant information from external knowledge bases and use them to augment the generation of tailored demonstrations. We leave these to future work.

\section{Supplementary Results}
\subsection{Combining \textsc{Self-Taught} with Other Zero-shot Prompting Methods}
We hereby present the full result of Table~\ref{tab:improvement}:
\begin{itemize}
    \item Table~\ref{tab:results_direct}: The results for \textsc{Self-Taught} that is combined with zero-shot Direct.
    \item Table~\ref{tab:results_ps}: The results for \textsc{Self-Taught} that is combined with zero-shot Plan-and-Solve (PS).
\end{itemize}

\subsection{Combining \textsc{Self-Taught} with a Smaller Open-source LLM}
We show the detailed results of Figure~\ref{fig:sllm} at:
\begin{itemize}
    \item Table \ref{tab:llama_qa}: \textsc{Self-Taught}'s performances when run with Llama-3.1-8B in question-answering.
    \item Table~\ref{tab:llama_ad}: \textsc{Self-Taught}'s performances when run with Llama-3.1-8B in the diagnosis of AD.
\end{itemize}

\section{Further Details on the Datasets}
\subsection{Question Answering of Diverse Domains}
\subsubsection{StrategyQA.}
This dataset contains questions that target multi-hop reasoning over a wide range of knowledge~\citep{geva2021did}. The main feature of this dataset is that the necessary knowledge required for solving the question is not explicitly stated in the question text
(\eg, ``Yes or No: \textit{Did Aristotle Use a Laptop?}'').
We adopted the test set (of 2,290 data) provided in BIG-bench.\footnote{\url{https://github.com/google/BIG-bench/tree/main}}


\subsubsection{ScienceQA.}
This dataset is proposed by \citet{lu2022learn}, questions are collected from elementary $\sim$ high school curricula, targeting 26 topics including math, language, geography, etc. We adopt the test set and exclude problems addressing the vision 
modality, \ie, images (the final test set has a size of 2,224.)
An example question is shown below:
\begin{quote}
Complete the statement. Hydrogen chloride is

(A) "a compound"

(B) "an elementary substance"
\end{quote}

\subsubsection{MedQA.}
This is a popular benchmark datasets in the medical domain curated by \citet{jin2021disease}. Since the questions are collected from national medical licensing exams in several countries (\ie, multi-lingual), we only adopt questions written in English. We use the test set with a size of 1,273.
An example is shown below:
\begin{quote}
    A 21-year-old sexually active male complains of fever, pain during urination, and inflammation and pain in the right knee. A culture of the joint fluid shows a bacteria that does not ferment maltose and has no polysaccharide capsule. The physician orders antibiotic therapy for the patient. The mechanism of action of action of the medication given blocks cell wall synthesis, which of the following was given?"

    (A) "Gentamicin"
    
    (B) "Ciprofloxacin"
    
    (C) "Ceftriaxone"
    
    (D) "Trimethoprim"
\end{quote}

\subsubsection{College-level QA of 6 domains.}
We include college-level problems of computer science (CS), medicine (Med), chemistry (Chem), math, physics (Phys), and biology (Bio) domains. The problems are collected from college textbooks and exams by \citet{hendrycks2020measuring} as a part of the MMLU (Massive Multitask Language Understanding) dataset.
The statistic is provided in Table~\ref{tab:dataset_college} and an example is shown below:
\begin{quote}
    Nitronyl nitroxides are stable radicals in which the unpaired electron is coupled to two equivalent nitrogen nuclei. How many lines will appear in the EPR spectrum of a solution of a rigid nitronyl nitroxide diradical with J $<<$ a?
    
    (A) 3 lines
    
    (B) 9 lines
    
    (C) 5 lines
    
    (D) 7 lines
\end{quote}

\begin{table}[h!]
\centering
\resizebox{0.9\columnwidth}{!}{
\begin{tabular}{l|cccccc}
\toprule
Domains &  CS &  Med &  Chem & Math & Phys & Bio\\
\midrule
Size & 100 & 173 & 100 & 100 & 102 & 144\\
\bottomrule
\end{tabular}%
}
\caption{Statistics of the college-level datasets (test sets).}
\label{tab:dataset_college}
\end{table}

\subsubsection{Professional-level QA of 4 domains.}
We include four datasets addressing professional-level knowledge of accounting (Acct), medicine (Med), psychology (Psych), and legal (Law) domains. The problems are mainly obtained from diverse licensing/bar exams of the corresponding profession by \citet{hendrycks2020measuring} as a part of the MMLU (Massive Multitask Language Understanding) dataset.
An example is shown below and the statistic is provided in Table~\ref{tab:dataset_pro}.
\begin{quote}
    An off-duty police officer was standing on a street corner waiting for a bus. A man came up from behind and stole the police officer's wallet from his pants pocket. As the man was running away with the wallet, the police officer pulled out his service revolver. The police officer yelled at the man to stop and then fired several shots in the man's direction. The police officer did not aim directly at the man but shot at the pavement intending to frighten him. One of the bullets ricocheted off the sidewalk and struck the man, killing him. The police officer is guilty of
    
    (A) assault with a deadly weapon. 
    
    (B) involuntary manslaughter.
    
    (C) voluntary manslaughter.
    
    (D) murder.
\end{quote}

\begin{table}[h!]
\centering
\resizebox{0.9\columnwidth}{!}{
\begin{tabular}{l|cccc}
\toprule
Domains &  Accounting &  Med &  Psychology & Law \\
\midrule
Size & 282 & 272 & 612 & 1534 \\
\bottomrule
\end{tabular}%
}
\caption{Statistics of the pro-level datasets (test sets).}
\label{tab:dataset_pro}
\end{table}

\subsection{AD Diagnosis with Real Patients}
\subsubsection{ADNI.}
The Alzheimer’s Disease Neuroimaging Initiative (ADNI)~\citep{jack2008alzheimer} is a long-term research project with the goal of addressing the diagnosis of AD. Data from ADNI has been widely used in prior works and significantly influenced the development of deep learning-based AD diagnosis~\citep{ebrahimi2020introducing,zhang2018multi, jang2022m3t, ong2023evidence}.

\subsubsection{AIBL.}
The Australian Imaging, Biomarker and Lifestyle Flagship Study of Ageing (AIBL)~\citep{ellis2009australian} is a project to investigate which biomarkers, cognitive characteristics, and health/lifestyle factors determine the subsequent progression of symptomatic AD. Data from AIBL is also one of the most widely used data for deep learning-based AD diagnosis~\citep{qiu2020development, zhu2021dual, jang2022m3t}.

\subsubsection{Features of both AD datasets.}
Each patient data from ADNI and AIBL has the following elements: (1) MRI scans of patients; (2) demographic information; (3) education level; (4) results from the mini-mental state examination; (5) the presence of APOE4 allele; (6) The ground-truth label of diagnosis.

\subsubsection{Textualized MRI data.}
Since this work focuses on mono-modal problem-solving, \ie, without considering the imaging modality, we incorporate the textualized ADNI and AIBL (only test sets) curated by \citet{kwon2024large}, where the MRI scans are transformed into textual descriptions in the form of EHR via an automatic process based on the structural features of brain regions.\footnote{\citet{kwon2024large} select 14 regions associated with AD: Hippocampus, Amygdala, Entorhinal, Parahippocampus, Medial Temporal Lobe, Fusiform, Precuneus, Superior Paretal, Lateral Ventricle, Frontal Lobe, Temporal Lobe, Parietal Lobe, Occipital Lobe, and Cerebral Cortex.}
We provide an example of such textualized data based on a patient from ADNI in Table~\ref{tab:appendix_AD_example}. The data derived from AIBL has the exact same format.

\begin{table}[h!]
\centering

\begin{tabular}{l}
\toprule
\textbf{Patient Description (EHR):}\\This patient is a \texttt{65}-year-old Male who has completed\\ \texttt{16} years of education and is \texttt{Married}.\\ The patient has a Mini-mental State Examination score\\of \texttt{26.0/30} and has \texttt{no} APOE4 gene.\\Also, based on their MRI scans:\\- This patient has \texttt{SEVERE} hippocampal atrophy.\\- This patient has \texttt{MILD}...\\ - This patient has \texttt{NO}...\\ ......\\\textbf{Diagnosis:}\\ \texttt{Alzheimer's Disease}\\
\bottomrule
\end{tabular}%

\caption{Partially masked example of ADNI data (one must be authorized by ADNI for full access). The \texttt{typewriter font} indicates values being inserted in the EHR template as mentioned in RQ2 result.}
\label{tab:appendix_AD_example}
\end{table}

\subsubsection{Statistics.}
The statistics of the test sets we used for our experiment is provided in Table~\ref{tab:dataset_statistics}.

\begin{table}[h!]
\centering
\resizebox{0.9\columnwidth}{!}{
\begin{tabular}{l|ccc}
\toprule
Diagnoses & \# AD & \# MCI & \# NC \\
\midrule
ADNI & 248 & 259 & 252 \\
AIBL & 130 & 158 & 140 \\
\bottomrule
\end{tabular}%
}
\caption{Statistics of the two AD datasets (test sets).}
\label{tab:dataset_statistics}
\end{table}

\subsubsection{Additional ethical statements.}
All data for AD diagnosis used in this work are approved by the Institutional
Review Board. They should not be shared without permission and
only be used by researchers authorized by ADNI and AIBL for research purposes. \textbf{Therefore, all patient information shown as examples is partially masked or omitted}.

\section{Demonstrations for Manual CoT and Few-shot Direct}
We provide the prompts for these baselines on our GitHub page.\footnote{\textbf{[link omitted during the review period]}} Here, we discuss how we acquire the few-shot demonstrations for them:
\subsection{Our Expert Annotation for MMLU Tasks.}
For college-level QA of 4 domains and profession-level QA of 6 domains from MMLU, we manually annotated the few-shot examples (\ie, CoT rationales) for Manual CoT with a group of experts of corresponding domains. The names of the experts are listed in Acknowledgements.
Note that because some of our experts do not speak English as their first language, generative AI may be applied during the annotation process purely for text translation/correction purposes.

\subsection{Demonstrations Annotated by Prior Work}
For StrategyQA, we adopt the human-annotated few-shot demonstration from \citet{wei2022chain}; For ScienceQA, we use the golden explanations provided in the original dataset as the CoT rationales in the few-shot demonstrations for Manual CoT; For MedQA, we use the human-annotated demonstrations provided by \citet{singhal2023large}; For ADNI and AIBL, we adopt the clinical CoT rationales provided by \citet{kwon2024large}.\footnote{\url{https://github.com/ktio89/ClinicalCoT}}. We also present all of them on our GitHub page.\footnote{\textbf{[link omitted during the review period]}}

\subsection{Demonstrations for Few-shot Direct}
For demonstrations in the few-shot Direct, we adopt the above-mentioned demonstrations and remove their rationale parts (\ie, only the problem text and the final answer are preserved).

\section{Codes and Prompts for \textsc{Self-Taught}}
\subsection{Codes for \textsc{Self-Taught}}
We provide the code for our proposed framework in our GitHub page.\footnote{\textbf{[link omitted during the review period]}}

\subsection{Prompts for \textsc{Self-Taught}}
We hereby provide the prompts for each of \textsc{Self-Taught}'s phases.
While the prompt structures for QA and AD diagnosis are exactly identical, we make small adjustments (\eg, [QUESTION] $\xrightarrow{}$ [PATIENT CASE]) due to the nature of the tasks: 
\begin{itemize}
    \item Figure \ref{fig:qa_prompt}: Prompts for \textsc{Self-Taught} in QA tasks.
    \item Figure \ref{fig:ad_prompt}: Prompts for \textsc{Self-Taught} in AD diagnosis.
\end{itemize}

\begin{table*}[t!]
\setlength{\tabcolsep}{3pt}
\centering
\small
\begin{tabular}{lccccc|cc||c}
\toprule

\textbf{Settings / Tasks}
 & StrategyQA & ScienceQA  & MedQA & COLLEGE & PRO & ADNI & AIBL & \textbf{Avg}\\ 
\midrule
Zero-shot Direct & 66.11 & 82.42 & 56.64 & 52.16 & 53.33 & 55.20 & 51.87 & 59.68 \\
\midrule
\textbf{\textsc{Self-Taught}} (based on Zero-shot Direct) & \textbf{70.22} & 84.40 & 57.42 & \textbf{55.59} & 54.55 & \textbf{59.68}& \textbf{52.34} & \textbf{62.03} (+2.35)\\
w/o Certainty Filtering & 69.35 & \textbf{85.75} & \textbf{57.66} & 54.66 & \textbf{54.88} & 55.47 & 50.70 & 61.21 (+1.53) \\

\bottomrule
\end{tabular}
\caption{Improvement brought by \textsc{Self-Taught} to zero-shot Direct.}
\label{tab:results_direct}
\end{table*}

\begin{table*}[t!]
\setlength{\tabcolsep}{3pt}
\centering
\small
\begin{tabular}{lccccc|cc||c}
\toprule

\textbf{Settings / Tasks}
 & StrategyQA & ScienceQA  & MedQA & COLLEGE & PRO & ADNI & AIBL & \textbf{Avg}\\ 
\midrule
Zero-shot PS & 70.39 & 87.72 & \textbf{66.93} & 63.84 & \textbf{58.30} & 56.62 & 55.14 & 65.56 \\
\midrule
\textbf{\textsc{Self-Taught}} (based on Zero-shot PS) & \textbf{71.67} & 87.99 & \textbf{66.93} & \textbf{65.23} & 57.52 & 59.03& \textbf{58.41} & \textbf{66.68} (+1.12) \\
w/o Certainty Filtering & 71.44 & \textbf{88.17} & 64.89 & 64.90 & 58.24 & \textbf{59.29}& 54.18 & 65.87 (+0.31) \\ 

\bottomrule
\end{tabular}
\caption{Improvement brought by \textsc{Self-Taught} to zero-shot Plan-and-Solve (PS).}
\label{tab:results_ps}
\end{table*}

\begin{table*}[t!]
\setlength{\tabcolsep}{3pt}
\centering
\small
\begin{tabular}{lccccccccccccc||c}
\toprule
& & & & \multicolumn{6}{c}{\textbf{MMLU: College Level}} & \multicolumn{4}{c}{\textbf{MMLU: Professional Level}} & \\
\cmidrule(lr){5-10} \cmidrule(lr){11-14}
\textbf{Methods / Tasks}
 & StrategyQA & ScienceQA  & MedQA & CS & Med & Chem & Math & Phys & Bio & Acct & Med & Psych & Law & \textbf{Avg} \\ 
 \midrule
\multicolumn{5}{l}{\textit{Zero-shot prompting / without real demonstrations:}}  \\ 
\midrule
Zero-shot Direct& 67.47 &75.22 &57.58 &47.00  & 67.74  & 50.00 &26.00 &50.98 &74.31 &47.16 &70.59 &61.44 &42.7 & 56.55\\

Zero-shot CoT & 72.20 &\textbf{92.48} &63.14 &\textbf{67.00} &\textbf{72.83} &\textbf{58.00} &53.00 &62.75 & \textbf{81.94}  & 58.51  & 83.46  &72.39 &50.07  & 68.29\\
Plan-and-Solve & 71.20&92.22&\textbf{67.56} & 62.00 & 68.21 & 54.00 & 44.00 & 65.69 & 78.47 & 56.70 & \textbf{81.60} & 67.97 & 49.47 & 66.08\\

\underline{\textbf{\textsc{Self-Taught}}}&  \textbf{73.95} & 90.20 & 67.24 & \textbf{67.00} & 69.36 & \textbf{58.00} & \textbf{57.00} & \textbf{71.57} & 79.71 & \textbf{58.87} & \textbf{81.60} & \textbf{74.51} & \textbf{52.22} &  \textbf{69.28} \\

\midrule
\midrule
\multicolumn{8}{l}{\textit{Oracles (demonstrations are made with real problems):}}
  \\
\midrule
Few-shot Direct & \underline{69.74} & \underline{82.33} & \underline{57.68} & \underline{46.39} & \underline{56.24} & \underline{43.30} & \underline{28.87} & \underline{48.04} & \underline{68.06}  & \underline{40.43}  & \underline{50.44} & \underline{61.27} & \underline{37.61}  & \underline{53.11}\\

Manual CoT & \underline{72.31} &\underline{88.62}&\underline{65.99} & \underline{61.86} & \underline{67.63} & \underline{46.39} & \underline{49.48} & \underline{66.83} & \underline{73.61} & \underline{50.71} & \underline{76.84} & \underline{66.83} & \underline{38.98} & \underline{63.54} \\

\bottomrule
\end{tabular}

\caption{Performances (question-answering) of Pareto efficient methods (presented in Figure~\ref{fig:cost}) when adopting a small open-source LLM, Llama-3.1-8B. \underline{Underlines} show oracles outperformed by ours. We report the accuracy.}
\label{tab:llama_qa}
\end{table*}

\begin{table*}[t!]
\centering
\setlength{\tabcolsep}{4pt}
\small
\begin{tabular}{lcccc}
\toprule
\textbf{Patients from:} &\multicolumn{4}{c}{\textbf{ADNI \& AIBL}} \\
\midrule
      \textbf{Methods / Metrics} & \textbf{F1} & \textbf{Precision} & \textbf{Recall} & \textbf{Accuracy} \\ 
\midrule 
Zero-shot Direct & 40.28 & 47.92 & 44.93 & 45.79\\
Zero-shot CoT & 52.10 & 52.90 & 54.96 & 53.75\\
Plan-and-Solve & 53.46 & \textbf{54.50} & \textbf{56.65} & 54.92\\

\textbf{\textsc{Self-Taught}} & \textbf{53.50} & 54.44 & 53.96 & \textbf{56.34}\\
\midrule \midrule
\multicolumn{5}{l}{\textit{Oracles (demonstrations are made with real problems):}}
\\ 
\midrule
Few-shot Direct & \underline{18.35} & \underline{38.20} & \underline{32.52} & \underline{35.47}\\
Manual CoT & 55.19 & 58.51 & 58.52 & 58.67\\

\bottomrule
\end{tabular}
\caption{Performances (AD diagnosis) of Pareto efficient methods (presented in Figure~\ref{fig:cost}) when adopting a small open-source LLM, Llama-3.1-8B. \underline{Underlines} show oracles outperformed by ours. Precision, recall, and F1 are weighted Avg of 3 diagnosis classes.}
\label{tab:llama_ad}
\end{table*}

\begin{figure*}[t!]
    \centering
    \includegraphics[width=1\textwidth]{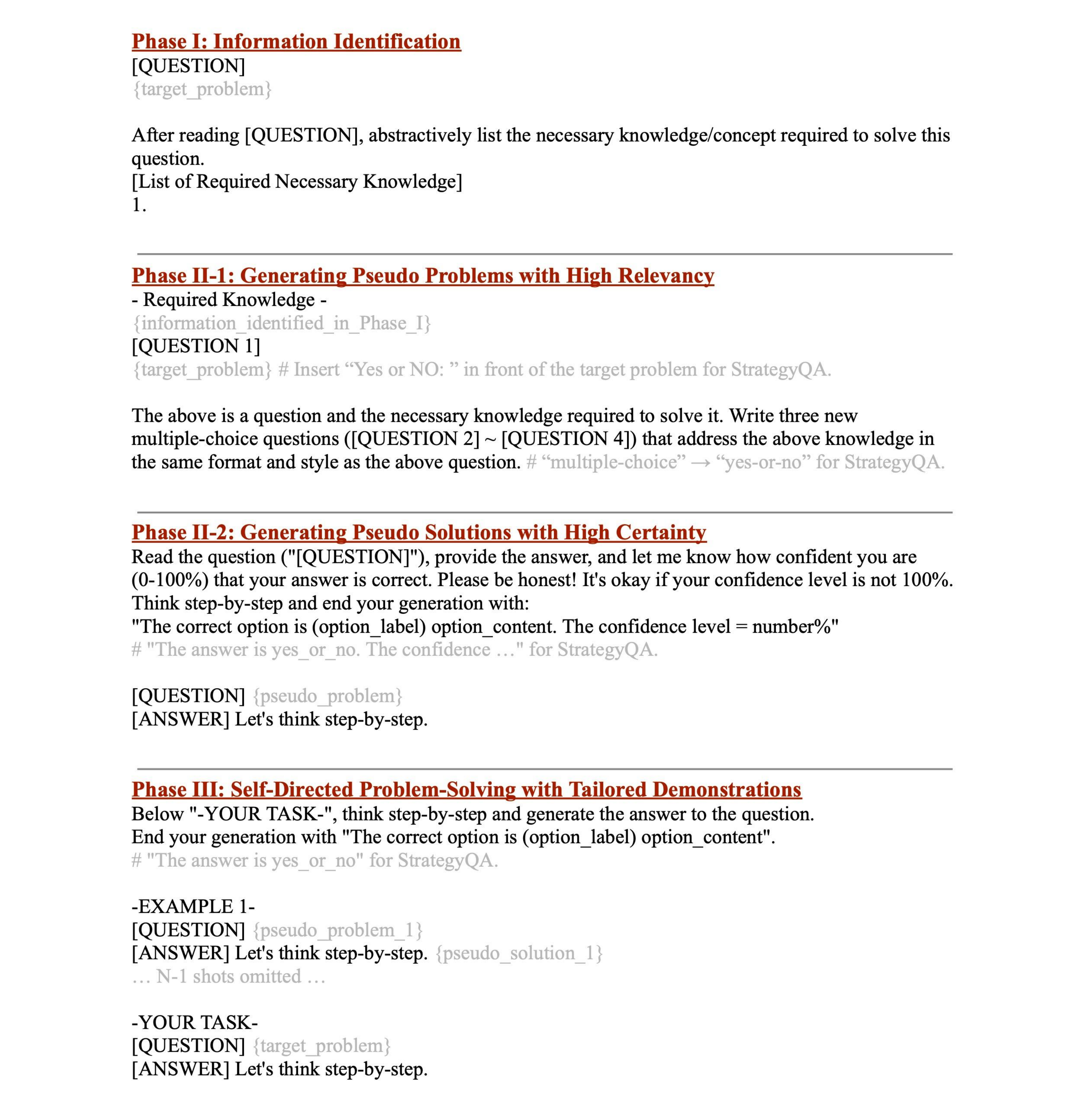}
    \caption{Prompts for all phases in \textsc{Self-Taught} for \textbf{question-answering} tasks. Gray comments (highlighted with ``\#'') show adjustments for different tasks, if any (not parts of the prompt). Phase I and II are performed completely under a zero-shot setting without any demonstrations. Phase II-2 is repeated if there is more than one pseudo problem to answer.
}
    \label{fig:qa_prompt}
\end{figure*}

\begin{figure*}[t!]
    \centering
    \includegraphics[width=1\textwidth]{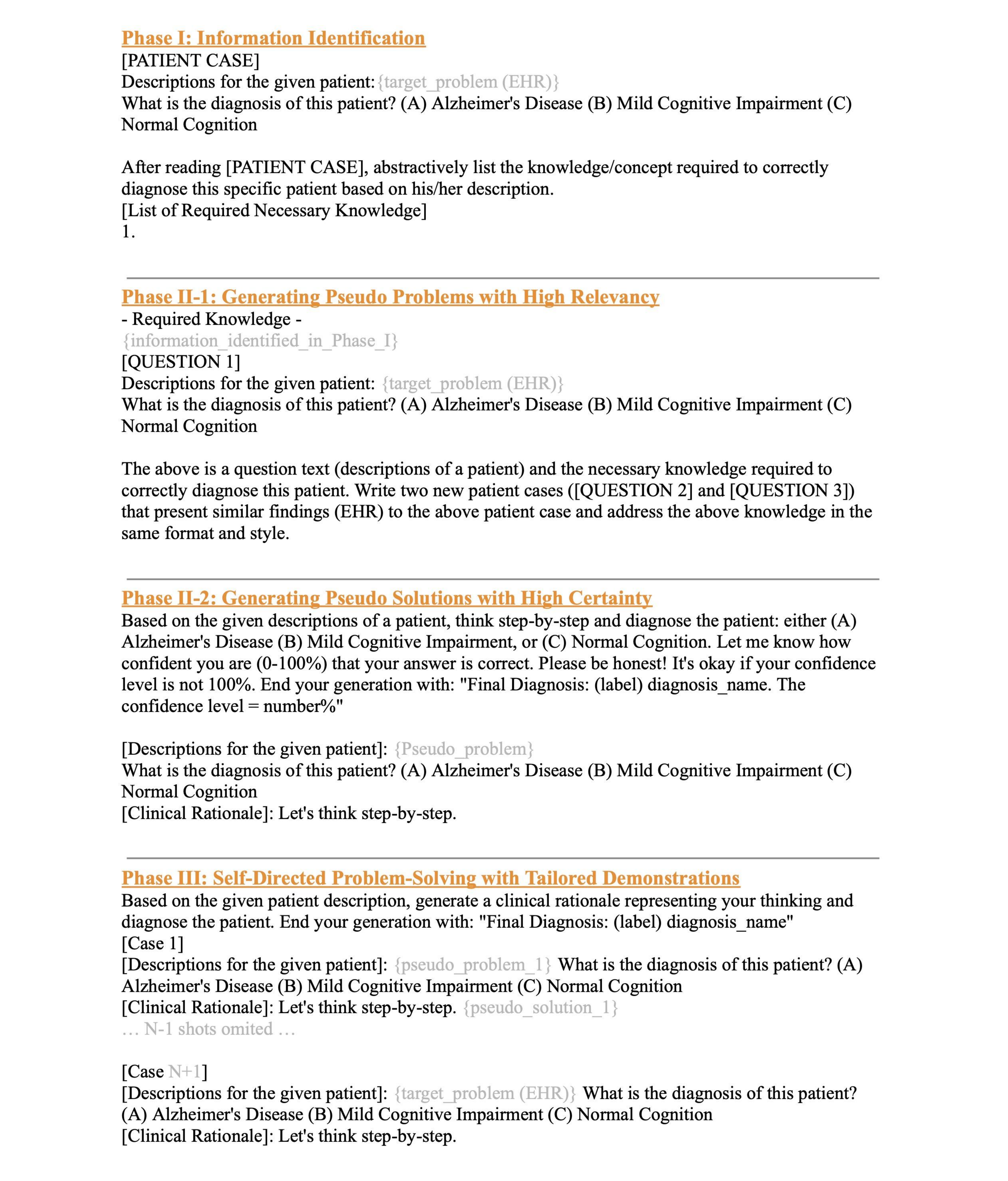}
    \caption{Prompts for all phases in \textsc{Self-Taught} for \textbf{AD diagnosis} tasks. Phase I and II are performed completely under a zero-shot setting without any demonstrations. Phase II-2 is repeated if there is more than one pseudo problem to answer.}
    \label{fig:ad_prompt}
\end{figure*}

\section{Empirical Examples of \textsc{Self-Taught}}
Besides Figure~\ref{fig:overview} and Figure~\ref{fig:case}, we provide several more examples of \textsc{Self-Taught}'s tailored demonstration in:
\begin{itemize}
    \item Figure~\ref{fig:example1} and~\ref{fig:example2}: We compare the tailored/relevant demonstrations in \textsc{Self-Taught} and Retrieval CoT.
    \item Figure~\ref{fig:example3} and \ref{fig:example4}: We show the generation of \textsc{Self-Taught}'s each phase.
\end{itemize}
We plan to present more examples on our GitHub page after the review period.

\section{Further Implementation Details}
As mentioned in Phase II-2, we empirically set $N$ to 3, $\lambda$ to 90, and $t$ to 5. We randomly select an $s$ with the highest $cs$ if we cannot obtain any $s$ with $cs \geq \lambda$ before the end of the $t$-th iteration.
The random seed for the random selection is 7 with the ``random'' module from Python.\footnote{\url{https://docs.python.org/3/library/random.html}}

\begin{figure*}[h!]
    \centering
    \includegraphics[width=0.9\textwidth]{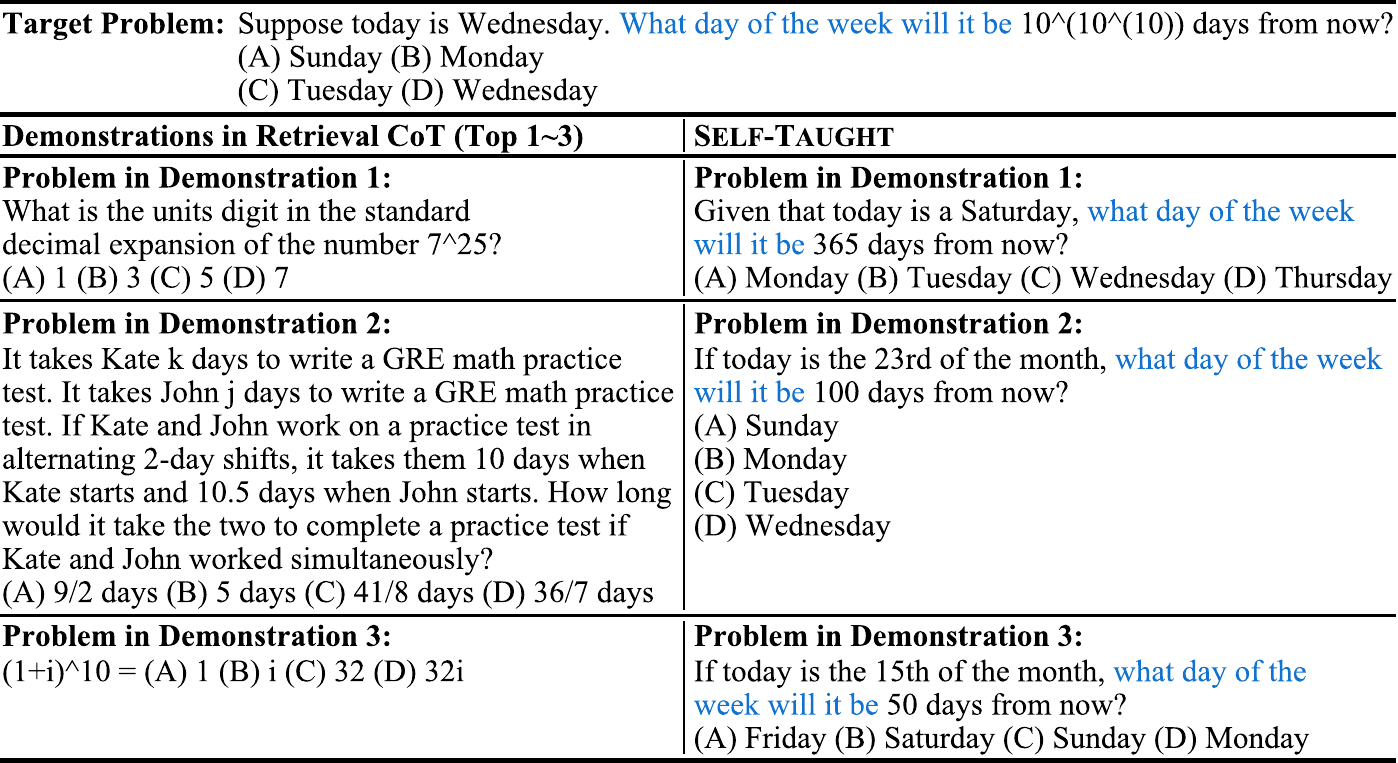}
    \caption{Demontrations in Retrieval CoT and \textsc{Self-Taught} \textbf{(1)}. Our method facilitates better relevant/tailored demonstrations than naively using in-domain corpora as the demonstration pool.}
    \label{fig:example1}
\end{figure*}

\begin{figure*}[h!]
    \centering
    \includegraphics[width=0.9\textwidth]{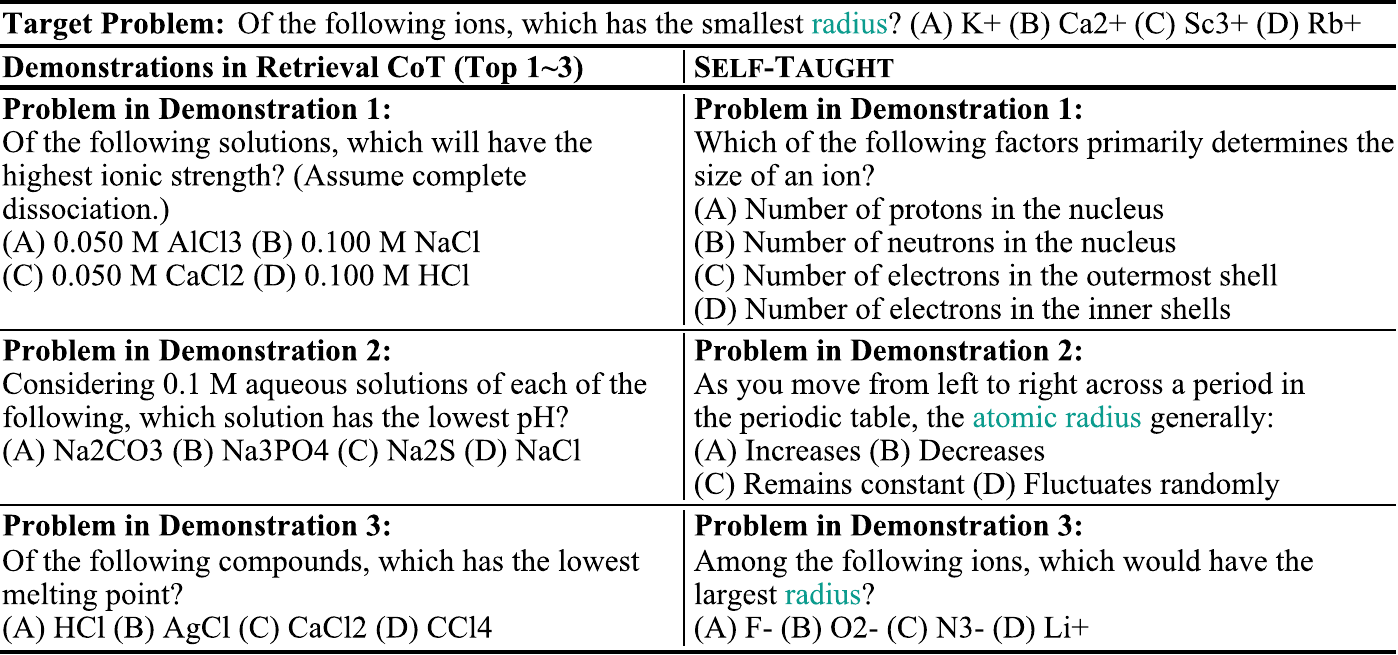}
    \caption{Demontrations in Retrieval CoT and \textsc{Self-Taught} \textbf{(2)}. Our method facilitates better relevant/tailored demonstrations than naively using in-domain corpora as the demonstration pool.}
    \label{fig:example2}
\end{figure*}

\begin{figure*}[h!]
    \centering
    \includegraphics[width=0.8\textwidth]{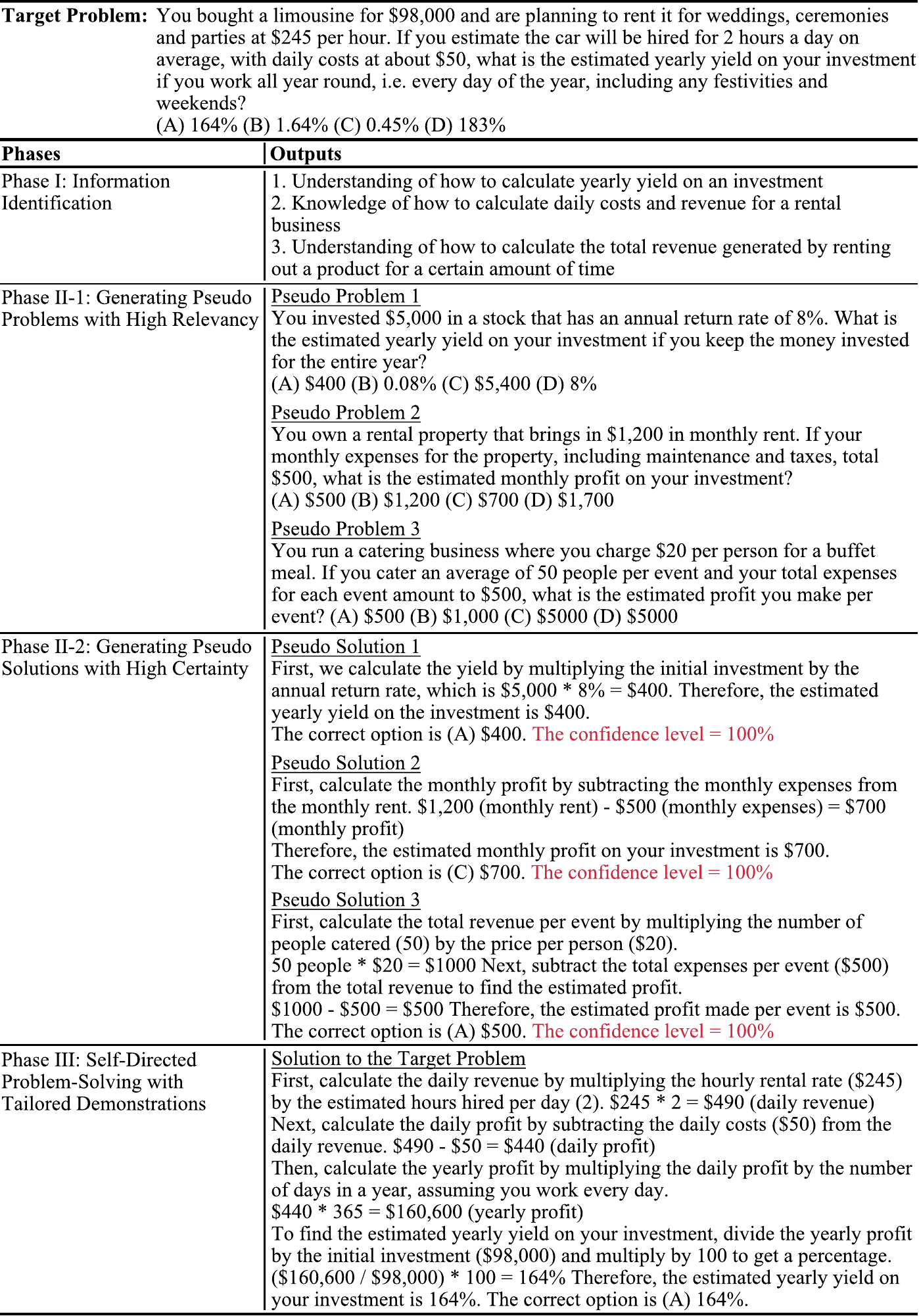}
    \caption{Examples of \textsc{Self-Taught}'s generation. Red texts are removed when used as pseudo demonstrations in Phase III.}
    \label{fig:example3}
\end{figure*}

\begin{figure*}[t!]
    \centering
    \includegraphics[width=0.75\textwidth]{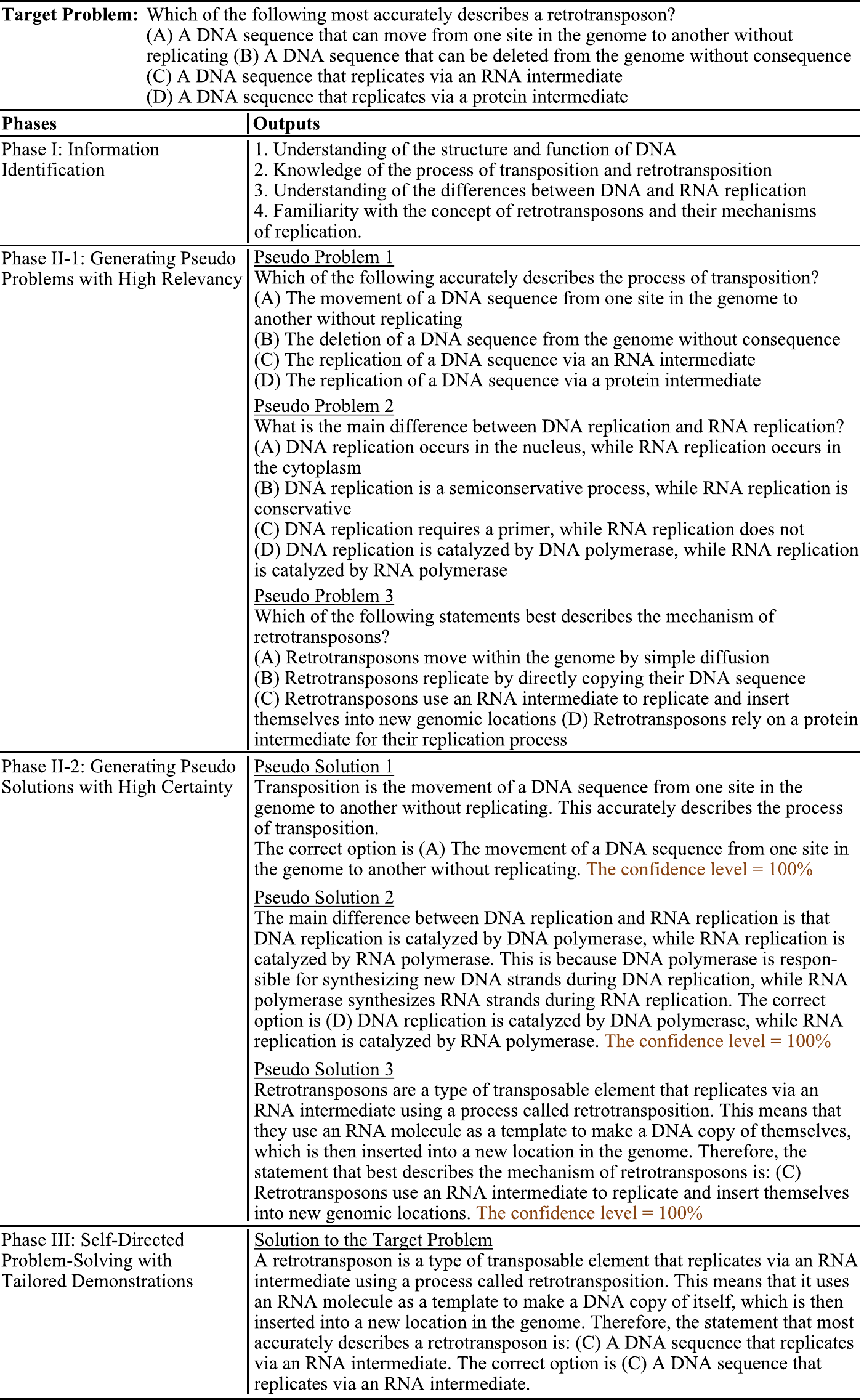}
    \caption{Examples of \textsc{Self-Taught}'s generation. Brown texts are removed when used as demonstrations in Phase III.}
    \label{fig:example4}
\end{figure*}

\end{document}